\documentclass{article}

\PassOptionsToPackage{colorlinks=true,linkcolor=black,citecolor=black,urlcolor=blue}{hyperref}
\usepackage[dblblindworkshop,final]{neurips_2026}
\workshoptitle{Foundation Models for Temporal Systems: From Forecasting to World Modeling}

\usepackage[utf8]{inputenc}
\usepackage[T1]{fontenc}
\usepackage{hyperref}
\hypersetup{
  pdftitle={Text-Conditioned Low-Rank Adaptation of Time-Series Foundation Models for Multimodal Forecasting},
  pdfauthor={Kunqi Li, Bin Liang, Zhidong Li},
  pdfsubject={Accepted at the NeurIPS 2026 Workshop on Foundation Models for Temporal Systems},
  pdfkeywords={time-series foundation models, multimodal forecasting, low-rank adaptation, text conditioning}
}
\usepackage{url}
\usepackage{booktabs}
\usepackage{amsmath,amsfonts,amssymb}
\usepackage{microtype}
\usepackage{xcolor}
\usepackage{graphicx}
\usepackage{multirow}
\usepackage{tabularx}
\usepackage{enumitem}
\usepackage{placeins}
\newcolumntype{L}[1]{>{\raggedright\arraybackslash}p{#1}}
\newcommand{\method}{C2LoRA-TS}
\newcommand{\gmmae}{GM-nMAE}

\title{Text-Conditioned Low-Rank Adaptation of Time-Series Foundation Models for Multimodal Forecasting}
\author{
  Kunqi Li \quad Bin Liang\thanks{Corresponding author: \texttt{bin.liang@uts.edu.au}.} \quad Zhidong Li\\
  University of Technology Sydney, Sydney, NSW, Australia\\
  {\small\texttt{kunqi.li@student.uts.edu.au \quad bin.liang@uts.edu.au \quad zhidong.li@uts.edu.au}}
}

\begin{document}
\maketitle

\begin{abstract}
Time-series foundation models (TSFMs) provide transferable numerical forecasting priors. However, forecast-relevant information may be conveyed in text without being reflected in the observed series. Many multimodal forecasting methods integrate such context through model inputs or intermediate representations. We investigate a complementary mechanism: text-conditioned parameter adaptation. We propose Context-to-LoRA for Time Series (\method{}), which encodes text and generates temporary Low-Rank Adaptation (LoRA) updates for selected weight matrices of a frozen TSFM. Both the numerical backbone and text encoder remain frozen, and the method exactly recovers the frozen TSFM output when context is absent. Evaluated on Time-MMD across nine domains and three TSFM architectures, \method{} achieves lower aggregate forecasting error than the frozen-TSFM and Static LoRA baselines for every architecture and seed. Replacing the instance-specific text with a single globally fixed context degrades performance for every architecture. These results support text-conditioned parameter generation as a viable strategy for multimodal forecasting with frozen TSFMs.
\end{abstract}

\section{Introduction}

Time-series foundation models (TSFMs) learn strong numerical forecasting
priors by pretraining on large, heterogeneous collections of time series
\citep{ansari2024chronos,das2024timesfm,liu2025moirai}. These priors transfer
across sampling frequencies and application domains, often without
task-specific training~\citep{liang2024Foundation}. However, the observed numerical history contains only
part of the information available at the forecast origin~\citep{jiang2025Multimodal}, particularly under non-stationary
dynamics where evolving trends can reduce the predictive sufficiency of
historical observations \citep{di2026TAPE}. Supply disruptions,
policy announcements, product launches, and public-health interventions may
alter a future trajectory before their effects become visible in the observed
series. Since such information is commonly expressed in language,
multimodal time-series forecasting requires an effective way to incorporate
textual context into a numerical forecaster.

Existing methods incorporate textual information into TSFMs in several ways.
UniCA maps heterogeneous covariates into homogeneous series representations
and fuses them with temporal representations
\citep{han2026unica}, whereas Aurora jointly pretrains a multimodal foundation
model over temporal and non-temporal inputs \citep{wu2026aurora}. Although architecturally different, both incorporate contextual information
through learned representations, consistent with common feature- and
representation-level fusion patterns in multimodal prediction
\citep{huang2026multimodalRealEstate}.

We instead investigate a complementary parameter-level mechanism in which text generates a temporary per-instance update to a frozen TSFM. This approach uses context to parameterize a temporary update for the current forecast:
\begin{equation}
    \widehat{y}_{T+1:T+H}
    =
    F_{\theta_0+\Delta\theta_\phi(C_T)}
    \left(y_{1:T}\right),
    \label{eq:parameter-conditioning}
\end{equation}
where $\theta_0$ denotes the frozen numerical TSFM and
$\Delta\theta_\phi(C_T)$ is the per-instance update produced from
textual context $C_T$ by a shared generator with parameters $\phi$. 

Low-Rank Adaptation (LoRA) supplies a structured form for this update
\citep{hu2022lora}. Text-to-LoRA~\citep{charakorn2025texttolora}, Doc-to-LoRA~\citep{charakorn2026doctolora}, and SHINE~\citep{liu2026shine} further show that
descriptions or documents can generate LoRA adapters without per-instance
optimization. We therefore ask whether textual context can generate useful forecast-specific low-rank adaptations for a frozen numerical TSFM.

\begin{figure}[t]
\centering
\includegraphics[width=\linewidth]{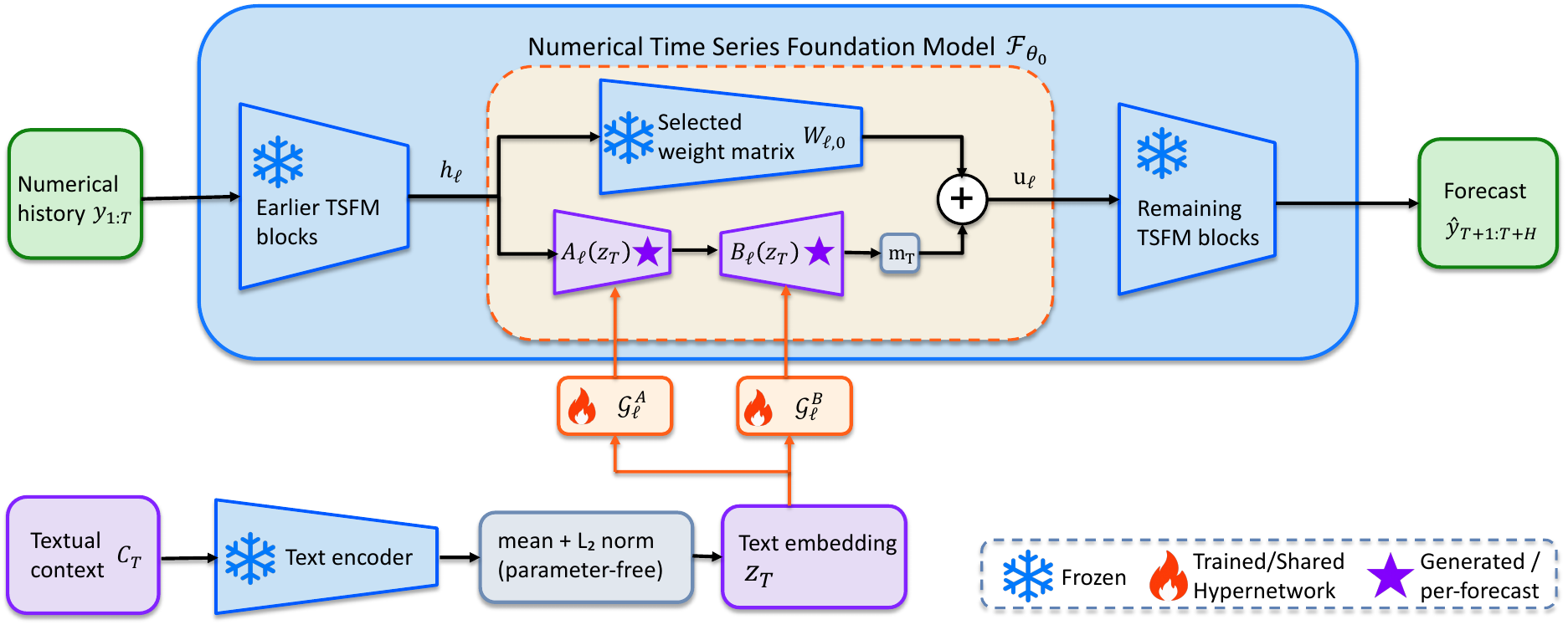}
\caption{Method overview. The numerical history is processed by a
frozen TSFM, while a frozen text encoder followed by normalized
mean pooling maps the textual context to a vector embedding $z_T$. For each selected frozen transformation $W_{\ell,0}$, hypernetworks $g_\ell^A$ and $g_\ell^B$
generate the low-rank adaptation factors $A_\ell(z_T)$ and $B_\ell(z_T)$. For each forecast,
the factor pair defines the rank-bounded update
$\Delta W_\ell(z_T)=m_TB_\ell(z_T)A_\ell(z_T)$, which is applied during the TSFM forward pass and vanishes when context is absent.}
\label{fig:method}
\end{figure}

Our implementation, Context-to-LoRA for Time Series (C2LoRA-TS), has three stages (Figure~\ref{fig:method}). First, a frozen sentence encoder embeds the available textual data, and a parameter-free normalized mean aggregates them into a forecast-level context vector. Second, for each selected TSFM weight matrix, two
learned factor-generation heads produce the factors of a low-rank update. Third, each factor pair defines a residual update that is applied to the corresponding TSFM transformation during the forecasting forward pass. During training, only the shared factor-generation heads are optimized using the forecasting objective; the TSFM and text encoder remain frozen. At inference, the factors are generated directly from context in a single forward pass, without per-forecast optimization.

We evaluate \method{} on Time-MMD across nine domains, three architecturally
distinct TSFMs, and three seeds. Our method outperforms both frozen
forecasting and Static LoRA for every backbone and seed. Replacing
instance-specific text with a fixed global context degrades performance across
all three backbones. Using a fixed context within each domain increases mean
error for all three backbones, with statistically supported differences for
two. Together, these results provide initial evidence that text-conditioned parameter
generation is a promising direction for multimodal forecasting with TSFMs.

\section{Text-Conditioned Low-Rank Adaptation}
\label{sec:method}

\paragraph{Problem formulation.}
At forecast time $T$, we observe the numerical history $y_{1:T}$ and textual
context $C_T$, and aim to predict the next $H$ values
$y_{T+1:T+H}$. Starting from a pretrained numerical TSFM
$F_{\theta_0}$ with frozen parameters $\theta_0$, we generate a temporary
parameter update for the current forecast:
\begin{equation}
\widehat y_{T+1:T+H}
=
F_{\theta_0+\Delta\theta_\phi(C_T)}(y_{1:T}).
\label{eq:forecast-formulation}
\end{equation}
Here $\Delta\theta_\phi(C_T)$ denotes the context-conditioned update produced
by a factor hypernetwork with trainable parameters $\phi$. The same
hypernetwork is learned across all training forecasts and generates the update
in a forward pass at inference time; no adapter is separately trained for an
individual forecast. 

\paragraph{Text representation.}
Suppose the textual context contains $J$ facts,
$C_T=\{c_1,\ldots,c_J\}$. A frozen text encoder $E$ maps each fact to an
embedding. For $J\geq 1$, we form a single representation of the complete
context as
\begin{equation}
e_j
=
\operatorname{norm}\!\left(E(c_j)\right),
\qquad
z_T
=
\operatorname{norm}\!\left(
    \frac{1}{J}\sum_{j=1}^{J}e_j
\right),
\label{eq:text-representation}
\end{equation}
where $\operatorname{norm}(a)=a/\lVert a\rVert_2$. Thus, $e_j$ is the
unit-normalized embedding of fact $j$, and
$z_T\in\mathbb R^{d_z}$ is the unit-normalized mean of all fact embeddings,
with $d_z$ denoting the text-embedding dimension. This aggregation is
parameter-free and treats the facts as an unordered set. When $J=0$, we set
$z_T=\mathbf{0}$.

\paragraph{Factor hypernetwork.}
Let $\ell$ index a selected frozen weight matrix
$W_{\ell,0}\in
\mathbb R^{d_\ell^{\mathrm{out}}\times d_\ell^{\mathrm{in}}}$.
Associated with this matrix are two trainable linear
factor-generation heads, $g_\ell^A$ and $g_\ell^B$. Given $z_T$, these heads
generate the entries of two low-rank adaptation factors:
\begin{equation}
\begin{aligned}
A_\ell(z_T)
&=
\operatorname{reshape}_{r\times d_\ell^{\mathrm{in}}}
\!\left(g_\ell^A(z_T)\right),\\[-1pt]
B_\ell(z_T)
&=
\operatorname{reshape}_{d_\ell^{\mathrm{out}}\times r}
\!\left(g_\ell^B(z_T)\right).
\end{aligned}
\label{eq:factor-generation}
\end{equation}
Here $r$ is the adaptation rank. The factor $A_\ell(z_T)$ projects the
transformation input into an $r$-dimensional adaptation space, while
$B_\ell(z_T)$ maps the resulting correction back to the output space.
The parameters of all factor-generation heads constitute $\phi$ and are
learned across forecasts.

To disable the generated update when text is unavailable, we define the fixed
availability mask
\begin{equation}
m_T=\mathbb{I}[J>0].
\label{eq:context-mask}
\end{equation}
This mask records only whether context is present and is not a learned gate.
For an input representation $h_\ell$, the generated factors define a residual transformation:
\begin{equation}
\begin{aligned}
\Delta W_\ell(z_T)
&=
m_TB_\ell(z_T)A_\ell(z_T),
&
\operatorname{rank}\!\left(\Delta W_\ell(z_T)\right)
&\leq r,\\[-1pt]
u_\ell
&=
W_{\ell,0}h_\ell+\Delta W_\ell(z_T)h_\ell,
\end{aligned}
\label{eq:adapted-forward}
\end{equation}
where $h_\ell$ is the input of the selected transformation and $u_\ell$ is the adapted output. The collection of
$\Delta W_\ell(z_T)$ over all selected transformations forms the full update
$\Delta\theta_\phi(C_T)$ in
Equation~\eqref{eq:forecast-formulation}.

\section{Experiments}
\label{sec:experiments}

\subsection{Experimental setup}

\paragraph{Dataset and context.}
We evaluate \method{} on Time-MMD, which pairs time series with textual
information across nine domains \citep{liu2024timemmd}. We use chronological
70/10/20 splits and evaluate all 4,522 test forecasts with a prediction horizon
of $H=12$. For each test forecast, we use the latest four nonempty
\texttt{search.fact} rows whose recorded end date precedes the first target
timestamp. Complete split statistics and context construction are provided in
Appendices~\ref{app:data}--\ref{app:text-construct}.

\paragraph{Backbones and comparisons.}
We study three architecturally distinct numerical TSFMs:
Chronos-Bolt-base, TimesFM~2.5, and Moirai~2.0. All backbone parameters remain
frozen. For each backbone, rank-eight updates are applied to the attention
output and feed-forward output transformations in two late blocks. We compare:

\begin{itemize}[leftmargin=*,nosep]
    \item \textbf{Frozen TSFM}, which uses only the pretrained numerical
    forecaster;

    \item \textbf{Static LoRA}, which learns one low-rank update shared by all
    forecasts; and

    \item \textbf{\method{}}, which generates a low-rank update from the
    textual context of each forecast.
\end{itemize}

\paragraph{Training and evaluation.}
Each trainable method is evaluated with three seeds. Learning rates are
selected on validation data, and model configurations are fixed before test
evaluation. We report the geometric mean across domains of MAE normalized by
the corresponding last-value forecast error (\gmmae{}; lower is better).
For context-replacement comparisons, we report paired moving-block 95\%
confidence intervals. Complete optimization, metric, and uncertainty
definitions appear in Appendices~\ref{app:training}--\ref{app:controls}.

To test whether \method{} uses variation in the textual context, we evaluate
each trained checkpoint under two replacements without retraining, following
recent concerns about semantic text sensitivity on Time-MMD
\citep{sridhar2026semantics}. The
\emph{global-fixed} condition supplies the same text to every forecast. The
\emph{domain-fixed} condition supplies one representative training context to
every forecast within the same domain. These controls preserve the trained
model while removing forecast-level variation from its textual input.

\subsection{Initial results}

\begin{figure*}[t]
\centering
\begin{minipage}[t]{0.49\textwidth}
\centering
\includegraphics[width=\linewidth]
{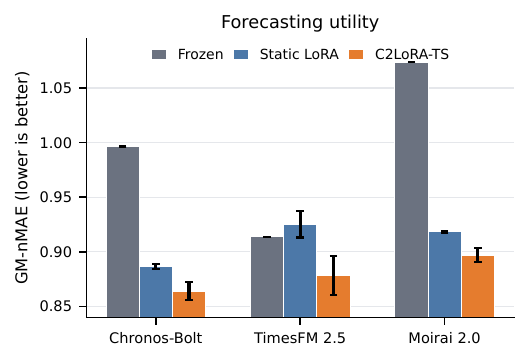}
\end{minipage}
\hfill
\begin{minipage}[t]{0.49\textwidth}
\centering
\includegraphics[width=\linewidth]
{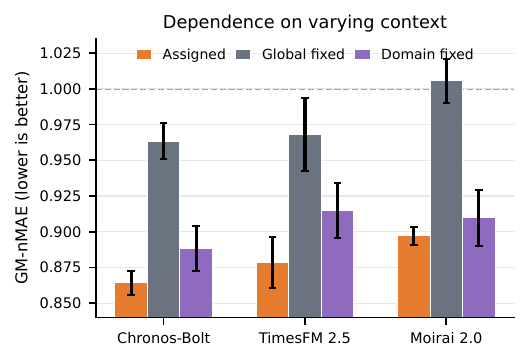}
\end{minipage}

\caption{
Initial Time-MMD results across three frozen TSFM backbones.
\textbf{Left:} forecasting utility of Frozen TSFM, Static LoRA, and
\method{}.
\textbf{Right:} the same \method{} checkpoints evaluated with assigned,
globally fixed, and domain-fixed context. Error bars show one standard
deviation across three seeds.
}
\label{fig:initial-results}
\end{figure*}

\paragraph{Forecasting performance.}
The left panel of Figure~\ref{fig:initial-results} compares \method{} with the
two numerical baselines. Our method achieves lower \gmmae{} than both the
frozen TSFM and Static LoRA for all three backbones and every training seed.
Relative to Static LoRA, \method{} reduces error by
2.52\% on Chronos-Bolt, 5.07\% on TimesFM, and 2.33\% on Moirai. 

\paragraph{Dependence on varying context.}
The right panel of Figure~\ref{fig:initial-results} evaluates the same trained
\method{} checkpoints under different context conditions. Assigned context also outperforms domain-fixed context on Chronos-Bolt and
TimesFM. The relative reductions are 2.69\% on Chronos-Bolt
(95\% CI: [1.07, 4.29]) and 3.99\% on TimesFM
(95\% CI: [1.91, 6.26]). The 1.39\% reduction on Moirai is not statistically
distinguishable from zero (95\% CI: [$-0.95$, 3.37]). Thus, forecast-varying
context provides information beyond a single globally fixed context for all three backbones
and beyond a fixed domain context for two of them. Exact numerical results are
reported in Appendix~\ref{app:context-results}, Table~\ref{tab:context-ladder}.

An additional replacement using an earlier, fact-disjoint context from the
same domain is not detectably different from assigned context; complete
results are provided in Appendix~\ref{app:context-results}. We therefore
interpret the present findings as initial evidence that varying textual
context can usefully generate TSFM parameter updates, rather than as
conclusive evidence of exact event attribution.

\FloatBarrier

\section{Conclusion}
\label{sec:conclusion}

\method{} outperforms both baselines for every backbone and seed, providing
initial evidence for the approach but not exact event attribution. Further work
is needed to assess its reliability across diverse forms of context and to
understand how textual information shapes the generated parameter updates.

\begingroup
\small
\bibliographystyle{plainnat}
\bibliography{references}
\endgroup

\clearpage
\appendix

\section{Method details and assumptions}
\label{app:method-details}

\subsection{Training objective}

Let $\phi$ collect the parameters of all factor-generation heads. We optimize
only $\phi$ using
\begin{equation}
\min_{\phi}\;
\mathbb E_{(y_{1:T},C_T,y_{T+1:T+H})\sim\mathcal D}
\left[
\mathcal L_{\mathrm{forecast}}\!\left(
F_{\theta_0+\Delta\theta_\phi(C_T)}(y_{1:T}),
y_{T+1:T+H}
\right)
\right].
\end{equation}
The numerical TSFM parameters $\theta_0$ and text encoder $E$ remain frozen
and in evaluation mode. 

\subsection{Text encoding}
\label{app:text-encoding}

We use the frozen \texttt{sentence-transformers/all-MiniLM-L6-v2} encoder. Each fact is
whitespace-normalized and prefixed with the fixed field label
\texttt{Search fact:} before encoding. Long facts are divided into
sentence-aligned chunks rather than silently truncated. Each chunk contains at
most 220 encoder tokens. Unit-normalized chunk embeddings are averaged by
encoder-token count and normalized again; the final mean across facts is also
normalized. The resulting vector $z_T$ has $d_z=384$ dimensions and no learned
pooling parameters. For empty context, $z_T=\mathbf 0$ and the fixed
availability mask is $m_T=0$.

\subsection{Static and generated LoRA}
\label{app:factor-details}

For
$W_{\ell,0}\in\mathbb R^{d_\ell^{\mathrm{out}}\times
d_\ell^{\mathrm{in}}}$, Static LoRA learns
$A_\ell\in\mathbb R^{r\times d_\ell^{\mathrm{in}}}$ and
$B_\ell\in\mathbb R^{d_\ell^{\mathrm{out}}\times r}$ once, so that
$\Delta W_\ell=B_\ell A_\ell$ is shared by every forecast. In \method{},
\begin{align}
g_\ell^A &: \mathbb R^{d_z}\rightarrow
\mathbb R^{r d_\ell^{\mathrm{in}}}, &
g_\ell^B &: \mathbb R^{d_z}\rightarrow
\mathbb R^{d_\ell^{\mathrm{out}}r},\\
\Delta W_\ell(z_T)&=m_TB_\ell(z_T)A_\ell(z_T).&&
\end{align}
Both heads are linear maps with bias. Their parameters are learned across the
training set, whereas $A_\ell(z_T)$ and $B_\ell(z_T)$ are generated anew for
each forecast.

The $A$ heads are initialized with small nonzero weights and biases; the $B$
heads are initialized to zero. Thus $\Delta W_\ell(z_T)=\mathbf 0$ at
initialization and the network exactly equals the frozen backbone.

\subsection{Empty-context identity and parameterization scope}

For $C_T=\varnothing$, the fixed mask gives $m_T=0$, hence
$\Delta\theta_\phi(C_T)=0$ and
$\widehat y_{T+1:T+H}=F_{\theta_0}(y_{1:T})$. This identity is architectural;
it is not learned from empty-context examples.

The factor heads' biases allow a component shared across nonempty contexts,
and the trained network may generate similar factors for different text.
Context dependence is therefore an empirical question, which motivates the
replacement tests in Section~\ref{sec:experiments}. Moreover, the rank bound
constrains each generated matrix update, not the total number of trainable head
parameters. We describe \method{} as low-rank adaptation but make no claim of
end-to-end parameter efficiency; generator capacity is reported in
Appendix~\ref{app:capacity}.

\subsection{Trainable capacity}
\label{app:capacity}

Across the four selected transformations, Static LoRA versus \method{} has
86,016 versus 33,116,160 trainable parameters for Chronos-Bolt; 81,920 versus
31,539,200 for TimesFM; and 34,816 versus 13,404,160 for Moirai. The generated
values per forecast are 86,016, 81,920, and 34,816, respectively. Thus, the
rank-matched Static LoRA comparison is structural rather than capacity matched;
we make no parameter-efficiency claim.

\section{Complete experimental protocol}

\subsection{Data and windows}
\label{app:data}

We use Time-MMD repository commit
\nolinkurl{00281e2d86058286d5548b15a7670e8eda57ef62}.

\begin{table}[h]
\centering
\caption{Complete Time-MMD window coverage.}
\small
\setlength{\tabcolsep}{5pt}
\begin{tabular}{llrrrr}
\toprule
Domain & Frequency & Lookback & Train & Validation & Test \\
\midrule
Agriculture & monthly & 8 & 353 & 42 & 96 \\
Climate & weekly & 8 & 871 & 116 & 244 \\
Economy & monthly & 8 & 293 & 34 & 79 \\
Energy & weekly & 36 & 1,088 & 151 & 314 \\
Environment & daily & 96 & 11,078 & 1,587 & 3,185 \\
Health\_US & weekly & 36 & 925 & 128 & 267 \\
Security & monthly & 8 & 197 & 20 & 51 \\
SocialGood & monthly & 8 & 627 & 82 & 166 \\
Traffic & monthly & 8 & 436 & 54 & 120 \\
\midrule
Total & --- & --- & 15,868 & 2,214 & 4,522 \\
\bottomrule
\end{tabular}
\end{table}

Rows are stably sorted by date. Split cutoffs are
$\lfloor0.7N_d\rfloor$ and $\lfloor0.8N_d\rfloor$ within domain $d$. A window
is retained only if all 12 target timestamps lie in its split and history and
target are finite; history may precede the split boundary. The univariate
target is the benchmark's designated \texttt{OT} column.

\subsection{Context construction}
\label{app:text-construct}

An eligible fact has nonempty \texttt{search.fact} text and
\texttt{end\_date} strictly earlier than the first target timestamp. Eligible
rows are stably ordered, and the latest four are returned chronologically.
Report predictions, future-context columns, numerical summaries, and upstream
no-information markers are prohibited. Every retained test case has four
eligible search rows under this rule. No \texttt{report.fact} field enters the
reported model.

\subsection{Backbones and adapted projections}
\label{app:backbones}

The frozen checkpoints and pinned revisions are:
\begin{itemize}[leftmargin=5mm,itemsep=0pt,topsep=2pt]
  \item Chronos-Bolt: \texttt{amazon/chronos-bolt-base}, revision\\
  \nolinkurl{5d9f166d69f47aef3401367a7b842e78fe97b121}.
  \item TimesFM~2.5: \texttt{google/timesfm-2.5-200m-transformers}, revision\\
  \nolinkurl{5a9806b9b291fad9233b5249d88263f1846304d3}.
  \item Moirai~2.0: \texttt{Salesforce/moirai-2.0-R-small}, revision\\
  \nolinkurl{30f43ff08c8494f4943ae1521e9d4e94a0fbb389}.
\end{itemize}

\begin{table}[h]
\centering
\caption{Functionally corresponding projections adapted in each frozen
backbone. All reported test models use rank eight.}
\small
\setlength{\tabcolsep}{4pt}
\begin{tabularx}{\linewidth}{L{0.20\linewidth}>{\raggedright\arraybackslash}X L{0.22\linewidth}}
\toprule
Backbone & Adapted transformations & Point forecast \\
\midrule
Chronos-Bolt-base & Blocks 10--11: attention \texttt{o}; feed-forward \texttt{wo} & Native median \\
TimesFM~2.5 & Layers 18--19: attention \texttt{o\_proj}; feed-forward \texttt{fc2} & Native mean \\
Moirai~2.0 & Layers 4--5: attention \texttt{out\_proj}; feed-forward \texttt{fc2} & Native 0.5 quantile \\
\bottomrule
\end{tabularx}
\end{table}

Chronos and TimesFM use their model-specific forecasting losses. For
evaluation, TimesFM uses its native \texttt{mean\_predictions} output, while
Chronos and Moirai use their native 0.5-quantile outputs. Moirai is trained
with mean pinball loss over nine quantiles. Histories shorter than Moirai's
96-step context are left-padded and marked by \texttt{past\_is\_pad}; no
observation is synthesized.

\subsection{Optimization and test discipline}
\label{app:training}

Seed 13 screens learning rates $\{10^{-4},3\!\times\!10^{-4},10^{-3}\}$ on
validation only. Static LoRA selects $10^{-3}$ for every backbone; \method{}
selects $10^{-3}$ for Chronos and $3\!\times\!10^{-4}$ for TimesFM and Moirai.
Seeds 37 and 71 reuse the selected rates. Training uses AdamW, weight decay
$10^{-4}$, batch size 64, clipping at 1.0, 5\% warmup and cosine decay, BF16,
and scaled dot-product attention. One balanced round contains a batch from each
domain. Training runs for at most 100 rounds, validates every 10 rounds on all
2,214 validation cases, and stops after at least four checks and three nonimproving
checks. Checkpoints are selected by validation \gmmae{}.

The Chronos protocol was locked before its test evaluation. TimesFM and Moirai
were prospectively locked replications after the Chronos result: they reused
the data, contexts, split, rank, projection-selection rule, optimizer rule, and initial
permutation. Their predictions were unknown until validation selection and
checkpoint locking were complete.

\subsection{Controls, metrics, and uncertainty}
\label{app:controls}

Assigned, Global fixed, Empty, and an initial within-domain mismatch belonged
to the initial evaluation. The initial mismatch changed every exact-context
hash but failed to ensure that donor text preceded its recipient forecast for
2,263/4,522 cases. It is therefore a mismatch stress test, not a causal control.
The Domain-fixed and Past-row controls were designed after the primary result,
locked, and applied once to frozen checkpoints without training or retuning.
Domain fixed uses the real training context nearest the domain's mean training
embedding. Past-row uses a different fact-disjoint same-domain context whose
recorded rows all precede the recipient target, with recent-row age matched
where possible.

For method $m$ and domain $d$, the primary score is
\begin{equation}
\operatorname{GM\text{-}nMAE}(m)=
\exp\!\left[\frac{1}{9}\sum_{d=1}^{9}
\log\!\left(\frac{\operatorname{MAE}_{m,d}}
{\operatorname{MAE}_{\mathrm{LV},d}}\right)\right].
\end{equation}
Here the last-value forecast repeats the last observation over the horizon.
Trainable-method scores average seeds 13, 37, and 71. For replacement condition
$c$, the reported point effect is
$100(1-\overline G_{\mathrm{Assigned}}/\overline G_c)$, where
$\overline G$ is the seed-mean GM-nMAE; positive values favor Assigned. Paired
intervals resample complete forecast trajectories within domain, recompute
domain ratios and the geometric mean, and then average seeds. Post-result
controls use 10,000 circular moving-block replicates with primary block length
12; lengths 1, 24, and 48 are sensitivity checks.

\FloatBarrier
\section{Seed-level forecasting results}
\label{app:initial-results}

\begin{table}[h]
\centering
\caption{Time-MMD \gmmae{} by seed; lower is better. Frozen is
deterministic, and trainable-method summaries report mean $\pm$ standard
deviation.}
\label{tab:appendix-initial-results}
\small
\setlength{\tabcolsep}{4.5pt}
\begin{tabular}{llrrrr}
\toprule
Backbone & Method & Seed 13 & Seed 37 & Seed 71 & Mean $\pm$ SD \\
\midrule
\multirow{3}{*}{Chronos-Bolt}
 & Frozen & .99645 & .99645 & .99645 & .99645 \\
 & Static LoRA & .88657 & .88419 & .88859 & .88645$\pm$.00220 \\
 & \method{} & .86153 & .87325 & .85756 & \textbf{.86411$\pm$.00816} \\
\midrule
\multirow{3}{*}{TimesFM~2.5}
 & Frozen & .91400 & .91400 & .91400 & .91400 \\
 & Static LoRA & .91123 & .93186 & .93275 & .92528$\pm$.01217 \\
 & \method{} & .87763 & .86072 & .89668 & \textbf{.87834$\pm$.01799} \\
\midrule
\multirow{3}{*}{Moirai~2.0}
 & Frozen & 1.07373 & 1.07373 & 1.07373 & 1.07373 \\
 & Static LoRA & .91928 & .91822 & .91773 & .91841$\pm$.00079 \\
 & \method{} & .90405 & .89380 & .89328 & \textbf{.89704$\pm$.00607} \\
\bottomrule
\end{tabular}
\end{table}

\section{Complete context ladder}
\label{app:context-results}

\begin{table}[h]
\centering
\caption{\method{} under every context condition, by seed. Initial mismatch is
pre-specified; Domain fixed and Past-row are post-result diagnostics.}
\label{tab:context-ladder}
\small
\setlength{\tabcolsep}{4pt}
\begin{tabular}{llrrrr}
\toprule
Backbone & Condition & Seed 13 & Seed 37 & Seed 71 & Mean $\pm$ SD \\
\midrule
\multirow{6}{*}{Chronos}
 & Assigned & .86153 & .87325 & .85756 & .86411$\pm$.00816 \\
 & Initial mismatch & .86880 & .87664 & .86167 & .86904$\pm$.00749 \\
 & Past-row & .85770 & .86665 & .85685 & .86040$\pm$.00543 \\
 & Domain fixed & .88990 & .90260 & .87138 & .88796$\pm$.01570 \\
 & Global fixed & .96908 & .97193 & .94915 & .96338$\pm$.01241 \\
 & Empty & .99645 & .99645 & .99645 & .99645 \\
\midrule
\multirow{6}{*}{TimesFM}
 & Assigned & .87763 & .86072 & .89668 & .87834$\pm$.01799 \\
 & Initial mismatch & .89056 & .87206 & .91047 & .89103$\pm$.01921 \\
 & Past-row & .87833 & .86146 & .89512 & .87830$\pm$.01683 \\
 & Domain fixed & .91736 & .89439 & .93282 & .91486$\pm$.01934 \\
 & Global fixed & .96001 & .94726 & .99678 & .96801$\pm$.02571 \\
 & Empty & .91400 & .91400 & .91400 & .91400 \\
\midrule
\multirow{6}{*}{Moirai}
 & Assigned & .90405 & .89380 & .89328 & .89704$\pm$.00607 \\
 & Initial mismatch & .91100 & .89801 & .90214 & .90371$\pm$.00664 \\
 & Past-row & .90457 & .89494 & .89416 & .89789$\pm$.00580 \\
 & Domain fixed & .92420 & .91758 & .88741 & .90973$\pm$.01961 \\
 & Global fixed & 1.00140 & .99294 & 1.02276 & 1.00570$\pm$.01537 \\
 & Empty & 1.07373 & 1.07373 & 1.07373 & 1.07373 \\
\bottomrule
\end{tabular}
\end{table}

\begin{figure}[h]
\centering
\includegraphics[width=.94\linewidth]{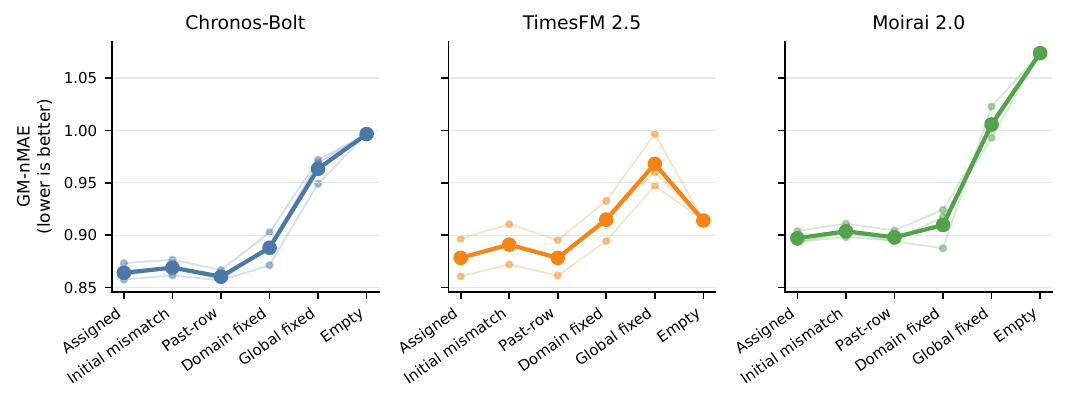}
\caption{Complete fixed-checkpoint context ladder. Thin lines denote seeds and
thick lines seed means.}
\end{figure}

For Assigned versus Past-row, the paired 95\% confidence intervals for the
relative effect are [$-1.02$, 0.14] on Chronos-Bolt, [$-0.63$, 0.71] on
TimesFM, and [$-0.75$, 0.94] on Moirai. Positive values favor Assigned; each
comparison contains 4,522 paired forecasts per backbone. All intervals use the
primary moving-block length of 12 and include zero; block lengths 1, 24, and 48
give the same conclusion.
\FloatBarrier

\section{Illustrative forecast trajectory and generating context}
\label{app:trajectory}

\begin{figure}[h]
\centering
\includegraphics[width=.96\linewidth]{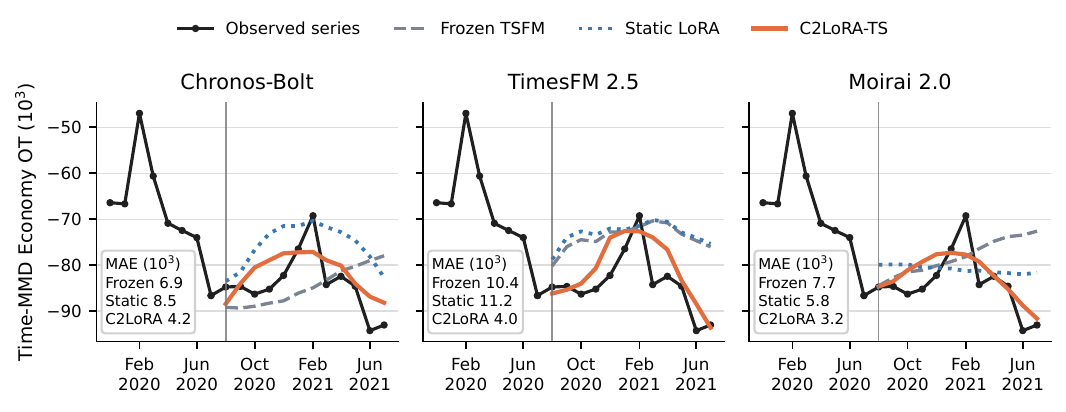}
\caption{Frozen, Static LoRA, and generated-LoRA trajectories for one
illustrative Economy forecast. The case was selected after inspecting outputs
and is not confirmatory evidence.}
\label{fig:trajectory-case}
\end{figure}

Figure~\ref{fig:trajectory-case} shows one Economy forecast whose target begins
1 August 2020. Every trajectory uses the same eight-month numerical history;
the four facts below generate the low-rank adaptation factors for \method{}.

\noindent\fcolorbox{black!22}{black!2}{%
\begin{minipage}{0.955\linewidth}
\small
\textbf{Exact normalized search-fact texts supplied to the factor generator}
\par\smallskip
\emph{Dates are source-row metadata and were not encoded.}
\begin{enumerate}[leftmargin=4.5mm,itemsep=2pt,topsep=3pt]
  \item \textbf{29 June--5 July 2020:} The United States entered into a
  recession in March 2020, as announced by the National Bureau of Economic
  Research (NBER) on June 8, 2020.
  \item \textbf{6--12 July 2020:} The U.S. economy experienced a decrease in
  Gross Domestic Product (GDP) in the first quarter of 2020, with all sectors
  contributing to the decrease, led by a decline in private inventory
  investment.
  \item \textbf{13--19 July 2020:} The U.S. economy is becoming more agile and
  innovative in response to continuing economic and environmental challenges.
  \item \textbf{20--26 July 2020:} The United States needs immigrants to
  strengthen its economy
\end{enumerate}
\end{minipage}}

\makeatletter
\setlength{\@fptop}{0pt}
\makeatother

\end{document}